\documentclass[10pt,twocolumn,letterpaper]{article}

\usepackage{wacv}
\usepackage{times}
\usepackage{epsfig}
\usepackage{graphicx}
\usepackage{amsmath}
\usepackage{amssymb}
\usepackage{xspace}
\usepackage{booktabs}
\usepackage{multirow}
\usepackage{subcaption}
\usepackage{url}
\usepackage{xcolor}
\usepackage[normalem]{ulem} 
\usepackage{verbatim}
\usepackage[font=small]{caption}
\usepackage{enumitem}
\usepackage{authblk} 
\usepackage{hyperref}
\hypersetup{
	colorlinks=true,
	linkcolor=red,
	filecolor=red,      
	urlcolor=red,
}

\wacvfinalcopy 

\def\wacvPaperID{370} 

\ifwacvfinal\fi
\begin{document}

\title{DeepSolarEye: Power Loss Prediction and Weakly Supervised Soiling Localization via Fully Convolutional Networks for Solar Panels}

\author[1]{Sachin Mehta}
\author[2]{Amar P. Azad}
\author[2]{Saneem A. Chemmengath}
\author[2]{Vikas Raykar}
\author[2]{Shivkumar Kalyanaraman}
\affil[1]{University of Washington, Seattle, WA, USA}
\affil[2]{IBM Research Lab, India \authorcr \small{Email:  {\tt\{sacmehta\}@uw.edu,  \{amarazad, saneem.cg, viraykar, shivkumar-k\}@in.ibm.com}}}

\maketitle
\ifwacvfinal\thispagestyle{empty}\fi

\begin{abstract}
The impact of soiling on solar panels is an important and well-studied problem in renewable energy sector. In this paper, we present the first convolutional neural network (CNN) based approach for solar panel soiling and defect analysis. Our approach takes an RGB image of solar panel and environmental factors as inputs to predict power loss, soiling localization, and soiling type. In computer vision, localization is a complex task which typically requires manually labeled training data such as bounding boxes or segmentation masks. Our proposed approach consists of specialized four stages which completely avoids localization ground truth and only needs panel images with power loss labels for training. The region of impact area obtained from the predicted localization masks are classified into soiling types using the \textit{webly supervised} learning. For improving localization capabilities of CNNs, we introduce a novel bi-directional input-aware fusion (BiDIAF) block that reinforces the input at different levels of CNN to learn input-specific feature maps. Our empirical study shows that BiDIAF improves the power loss prediction accuracy by about 3\%  and localization accuracy by about 4\%. Our end-to-end model yields further improvement of about 24\% on localization when learned in a weakly supervised manner. Our approach is generalizable and showed promising results on web crawled solar panel images. Our system has a frame rate of 22 fps (including all steps) on a NVIDIA TitanX GPU. Additionally, we collected first of it's kind dataset for solar panel image analysis consisting 45,000+ images.
\end{abstract}

\vspace{-2mm}
\section{Introduction}
\vspace{-2mm}
The surge in solar photovoltaic (PV) based renewable energy in recent years has revolutionized the energy sector across the globe by greatly reducing the energy cost \cite{usa2020}. Growing number of large- and mid-sized solar farms often face operations and maintenance challenges. Environment induced soiling on solar panel (accumulation of dust, pollen, leaves, bird drop, snail trail, and snow) and defects,  such as cracks, hamper the power generation at large \cite{DustReview,MANI20103124,SoilingPwrLoss,Zapata2015}.  Automatic visual inspection-based solutions can play a vital role in efficient solar farm operations, maintenance, and asset warranty.

\begin{figure}[!t]
\centering
\setlength{\belowcaptionskip}{-4mm}
\includegraphics[width=0.9\columnwidth]{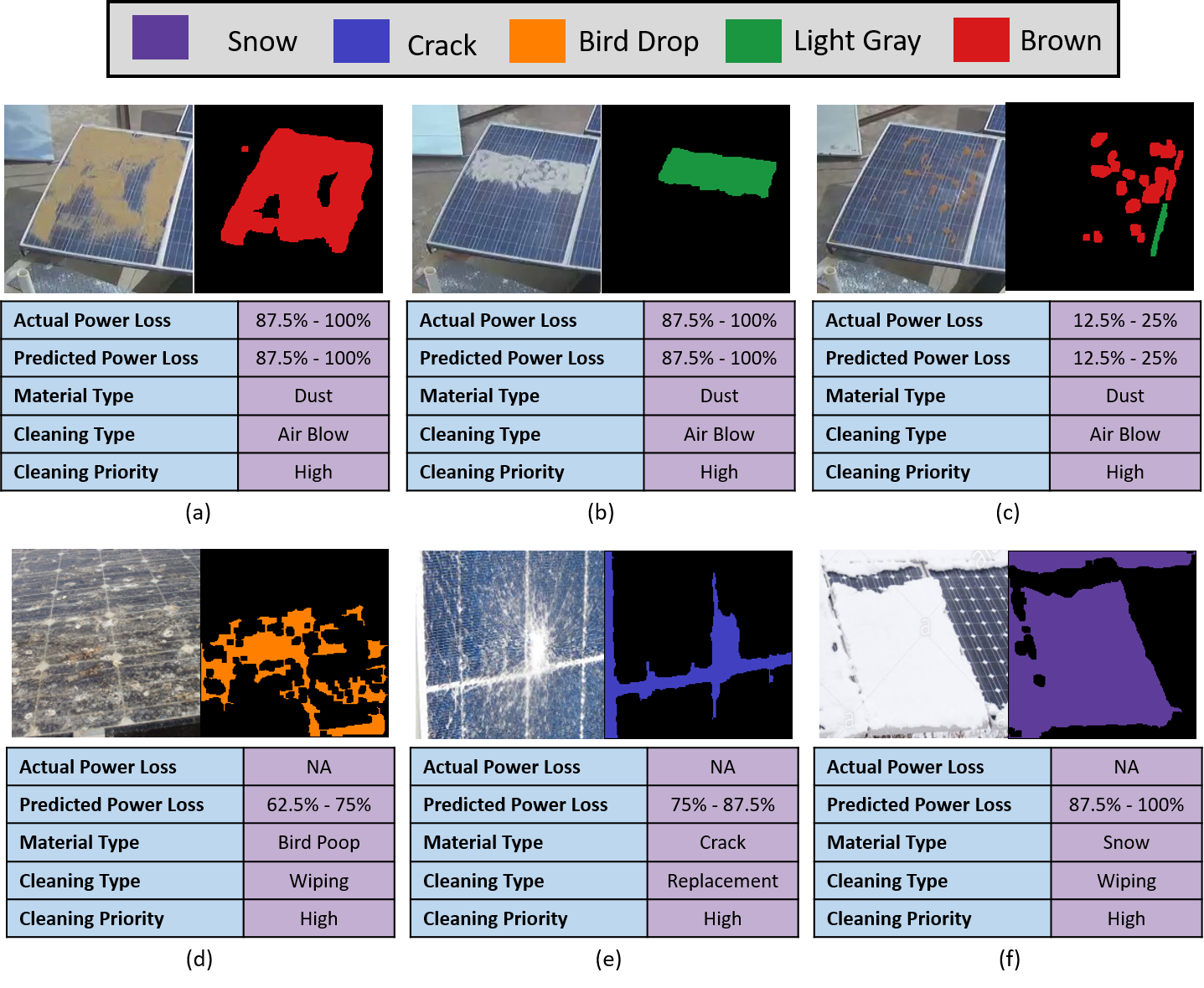}
\caption{Six images depicting the performance of our method. First three images are from our dataset while the remaining three are downloaded from the Internet. Our method can efficiently localize PV soiling and type, even in the wild. See \textbf{Appendix \ref{sec:moreRes}} for more images. Best viewed in color.}
\label{fig:siximages}
\end{figure}

The type of soiling or defect on the panel can be instantly and effortlessly recognized by merely looking at the PV panel image. However, to analyze the impact of soiling or defect on the performance of the solar panel, detailed information, such as soiling amount and coverage, type of the material, and location on the panel, is required. This information is not only useful for estimating the impact on the performance of solar panel, but also helpful for recommending corrective measures; which together are critical for efficient solar farm maintenance. As an example, \textit{wiping} is a more appropriate cleaning action when solar panel is covered with bird drop rather than air blow which is effective for cleaning dust (see Figure \ref{fig:siximages}(c) and (d)). This decision is likely due to the nature of the material, e.g., bird drop has sticky and oily composition than dust.

\begin{figure}[t!]
\centering
\setlength{\belowcaptionskip}{-4mm}
\includegraphics[width=1.0\columnwidth]{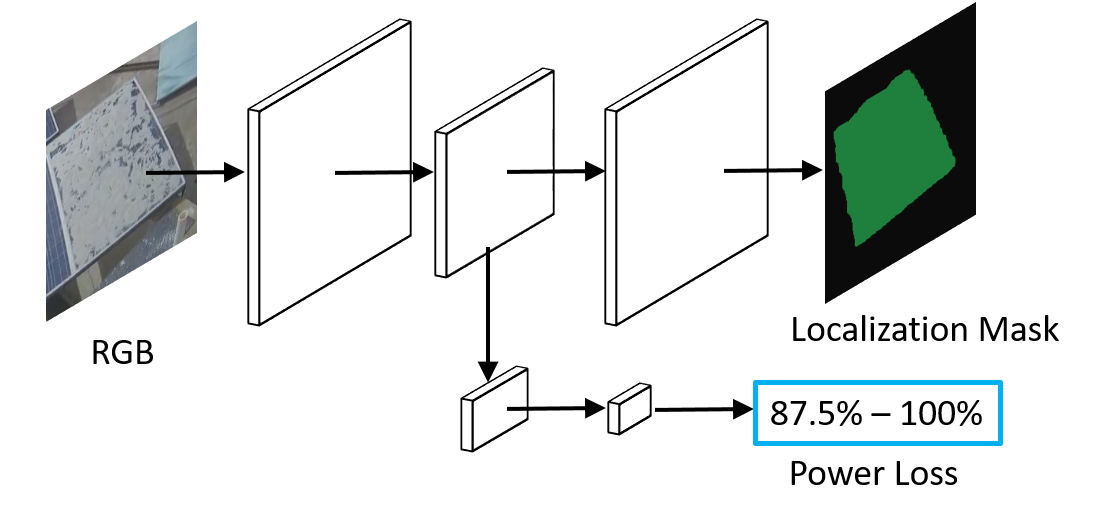}
\caption{Overview of our method, DeepSolarEye, that predicts impact on the power loss and the soiling area simultaneously.}
\label{fig:IMCNN_images}
\end{figure}
Besides soiling coverage and type, soiling location also plays a vital role in impact prediction due to the physics of cell connections in a panel.  As an example, dust cover on panels in Figure \ref{fig:siximages}(b) is much less than that in Figure \ref{fig:siximages}(a), however, power loss is similarly high in both the cases. 
By design, solar panels have cells in series in vertical columns and one bypass diode per two columns. Therefore, even if one cell is fully covered by dust, it can block the current for the entire bypass diode segment; thus resulting in high power loss.

We aim to develop a model that provides information about the soiling and defects on the panel using RGB image of a solar panel. Convolutional neural networks (CNNs) have performed well in visual recognition tasks such as object detection and image segmentation. Yet, application of CNNs in solar farm monitoring and management still remains a challenge; primarily due to lack of labeled datasets. 

In this paper, we present a novel end-to-end fully convolutional neural network, DeepSolarEye, that simultaneously predicts the power loss and localizes soiling area from an image of a solar panel.  A simplified overview of DeepSolarEye is presented in Figure \ref{fig:IMCNN_images}. Ideally, we need power loss and localization mask labels to train such a model. In our approach, we bypass the explicit localization mask requirement. We use power loss (or classification label) as weak supervision for generating the localization masks. 

The main contributions of our paper are: 
\begin{itemize}[itemsep=0pt]
\item[{(1)}] {\it \textbf{CNNs for solar panel analysis:}} We adapt existing CNNs for a new domain of solar panel soiling and defect analysis. To the best of our knowledge, ours is the first CNN-based approach for this task. 
\item[{(2)}] {\it \textbf{Weakly supervised learning:}} DeepSolarEye consists of four steps: (a) train a CNN-based classification network, ImpactNet, for predicting the power loss, (b) create a candidate soiling mask using a  pyramid-based approach from the classification network, (c) train a multi-task network, Mask FCNN, for simultaneously predicting the power loss and localizing the soiling area using the generated candidate mask, and (d) predict the soiling category using webly supervised neural network, WebNN. 
\item[{(3)}] {\it \textbf{ BiDIAF for accurate localization:}} CNNs learn features at multiple spatial levels by performing convolution and down-sampling operations. However, feature maps learned by CNNs do not encode localization information explicitly; thereby hindering in localizing the impact area. Localization further becomes difficult as spatial information is lost due to these operations. To overcome these challenges, we introduce a novel convolutional unit, a bi-directional input-aware fusion (BiDIAF), that reinforces the input at different layers of CNNs to learn input-specific feature maps. We show that BiDIAF  improves the localization capabilities of CNNs largely. 
\item[{(4)}] {\it \textbf{ Dataset:}} We created a first of its-kind-dataset for solar panel image analysis, comprising of 45,754 solar panel images with labels of power loss and solar irradiance, as well as timestamps. DeepSolarEye achieves a classification accuracy of 83.32\% and localization Jaccard index (weakly supervised) of 66\% on this dataset. Further, our experimental results suggest that DeepSolarEye learns generalizable representations of different soiling types and able to identify and localize the soiling on solar panels, even in the wild. For example, crack (Figure\ref{fig:siximages}(e)) and snow (Figure\ref{fig:siximages}(f)) were identified and localized despite being not present in our dataset.
\end{itemize}

\vspace{-2mm}
\section{Related Work}
\vspace{-2mm}
\paragraph{Image-based solutions for solar panel analysis:} There exist some work for analyzing different soiling types \cite{aghaei2015ir, yap2015quantitative,dotenco2016automatic, Aghaei2016}. These methods take RGB or IR\footnote{IR-based systems exploit the overheating phenomenon while RGB-based systems exploit the color information to locate the hot-spots .} image as an input and applies a traditional image processing algorithm, such as histogram matching, filtering, and color-space conversion. The output of the image processing algorithm is then thresholded to locate the impact area. As these methods are threshold-bound, they are not scalable to identify different types of soiling. Further, these methods are not able to capture the complex relationships between different factors, such as particle size, thickness, and coverage along with environmental factors (e.g. solar irradiance and humidity), that are required for analyzing the impact on power loss. 

\vspace{-4mm}
\paragraph{CNN for visual recognition tasks:} Convolutional Neural Networks (CNNs) are the state-of-the-art methods for image classification \cite{simonyan2014very,szegedy2015going,ResNet,huang2017densely}. Recent classification architectures have explored different types of connectivity patterns, such as bypass connections in \cite{ResNet} and dense connections in \cite{huang2017densely}, to improve the information flow inside the network; thereby enabling end-to-end training of very deep CNNs. Further, these classification networks have been used as the base feature extractors for several visual recognition tasks including object detection \cite{ren2015faster} and segmentation \cite{chen2016deeplab}.

Region-based CNNs or R-CNNs \cite{girshick2014rich} have proven to be effective for both detection and segmentation tasks . However, the accuracy of R-CNNs is dependent on the region proposal method. Unlike R-CNNs, fully convolutional networks (FCN) have gained attention as they enable end-to-end training and are fast (e.g. \cite{ren2015faster, badrinarayanan2017segnet}). The features learned by the classification network at lower spatial resolution (say $7\times 7$) are coarse and leads to coarse output (e.g. segmentation masks of FCN-32s \cite{shelhamer2017fully}). To address this limitation, several techniques have been proposed such as fully convolutional region proposal networks (e.g. \cite{ren2015faster}), skip-connections (e.g. \cite{shelhamer2017fully, ronneberger2015u}), deconvolutional networks (e.g. \cite{badrinarayanan2017segnet}), dilated convolutions (e.g. \cite{chen2016deeplab} \cite{yu2015multi}), and multiple-input networks (e.g. \cite{lin2016refinenet}).

Several FCN-based supervised object classification and localization networks exists in literature. Yet, extending these approaches to new domains is challenging; primarily due to the lack of large labeled datasets. In this paper, we propose a method for PV solar image analysis to address these challenges. Our method consists of novel and carefully designed components that allows extending the prior work on image classification and localization on a new dataset \textit{without} any manually labeled localization data. 

\vspace{-2mm}
\section{Dataset}
\vspace{-2mm}
We create a first-of-its-kind dataset\footnote{More details about the project can be found here: \url{https://deep-solar-eye.github.io/}.}, PV-Net, comprising of 45,754 images of solar panels with power loss labels. Our experimental setup consists of two identical solar panels, which are kept side by side with an RGB camera facing them. Soiling experiments were conducted on the first panel (close to the camera) while the other panel was used for reference. Images were captured at every 5 seconds and power generated by the panels was recorded. Soiling impact is reported as the \textit{percentage power loss} with respect to the reference panel. We will be using soiling impact and power loss interchangeably in this paper.

Our data recording methodology was aimed to capture various types of soiling and their impact on PV panel. For this, we exposed the panel to different types of soiling in terms of color (red, brown, and gray), particle size (sand, dust, and talcum powder), and thickness under natural environmental conditions \cite{DustReview}. Some of the thick patches correspond to high power losses as much as 90\%. The data set was collected for about a month and was enriched with large variations in soiling due to both experimental (e.g. dust with varying thickness, blob sizes, and patches) and natural means (e.g. wind and precipitation). Besides power loss corresponding to the soiled panel, our data set also contains information about environmental factors (solar irradiance and timestamps).

\vspace{-2mm}
\section{DeepSolarEye}
\begin{figure*}[ht]
\centering
\includegraphics[width=1.8\columnwidth]{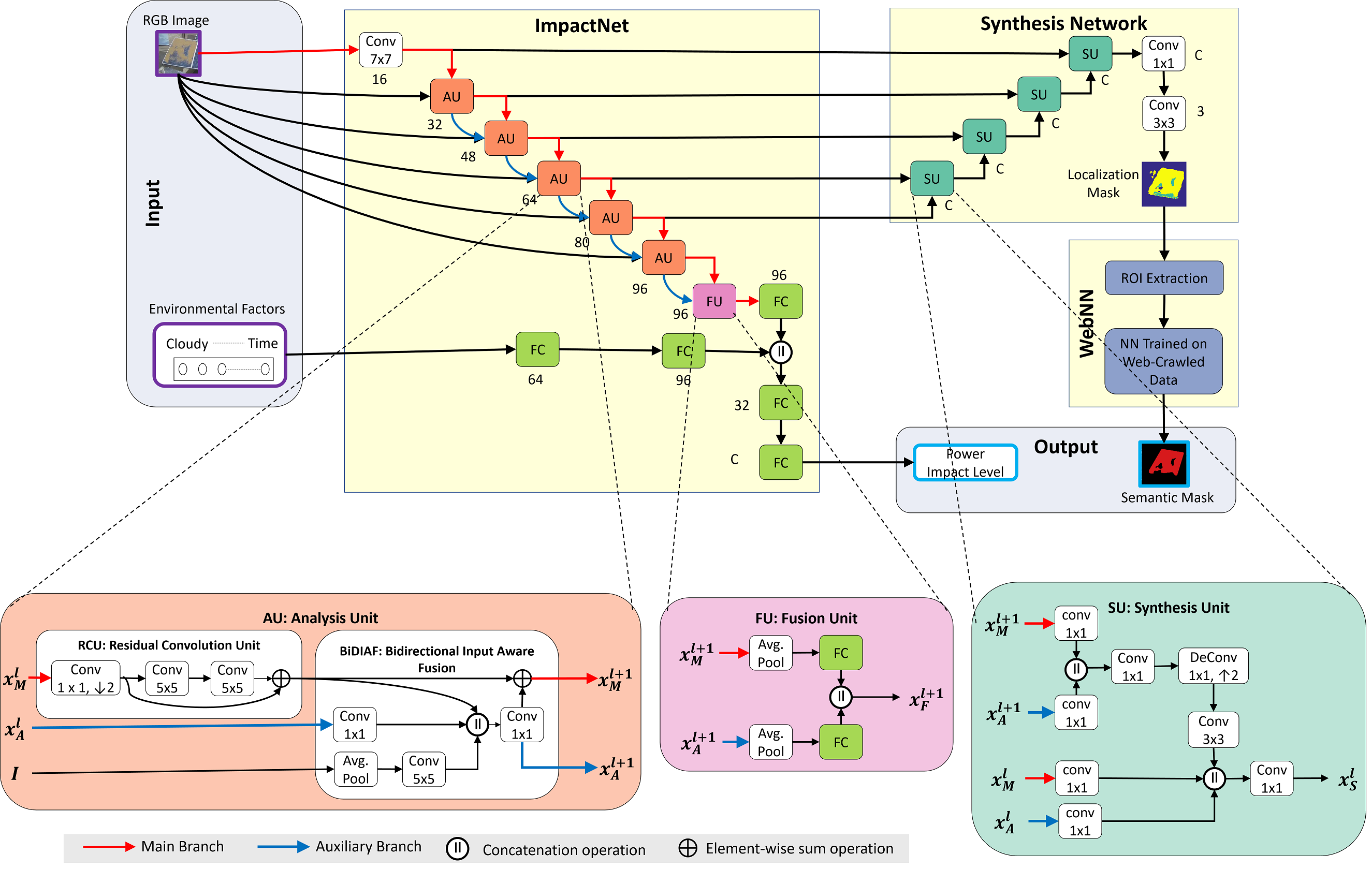}
\setlength{\belowcaptionskip}{-4mm}
\caption{DeepSolarEye: An end-to-end system for predicting the soiling impact, the soiling localization, and the soiling type simultaneously. Number of feature maps used by each block are reported next to it. Note that $1^{\textrm{st}}$ AU doesn't have $x^l_A$. Best viewed in color.}
\label{fig:resnetPP}
\end{figure*}

\vspace{-2mm}
We propose an end-to-end system, DeepSolarEye, which is based on fully convolutional networks. The architecture of DeepSolarEye is visualized in Figure \ref{fig:resnetPP}. The input to our system is an RGB image with environmental factors (such as solar irradiance and timestamps), while outputs of our system are: (1) soiling impact, (2) soiling localization, and (3) soiling  category. We propose a four-step approach that enables the training of DeepSolarEye: (1) train a CNN-based classification network, ImpactNet, for predicting soiling impact, (2) create candidate masks by aggregating the feature maps learned by classification network at different spatial-levels using pyramid-based approach, (3) train a multi-task network, Mask FCNN, for simultaneously predicting the soiling impact and soiling localization mask, and (4) predict the category of soiling using webly supervised neural network, WebNN. These steps are discussed below.
\begin{figure}[t!]
\centering
\setlength{\belowcaptionskip}{-4mm}
\includegraphics[width=\columnwidth]{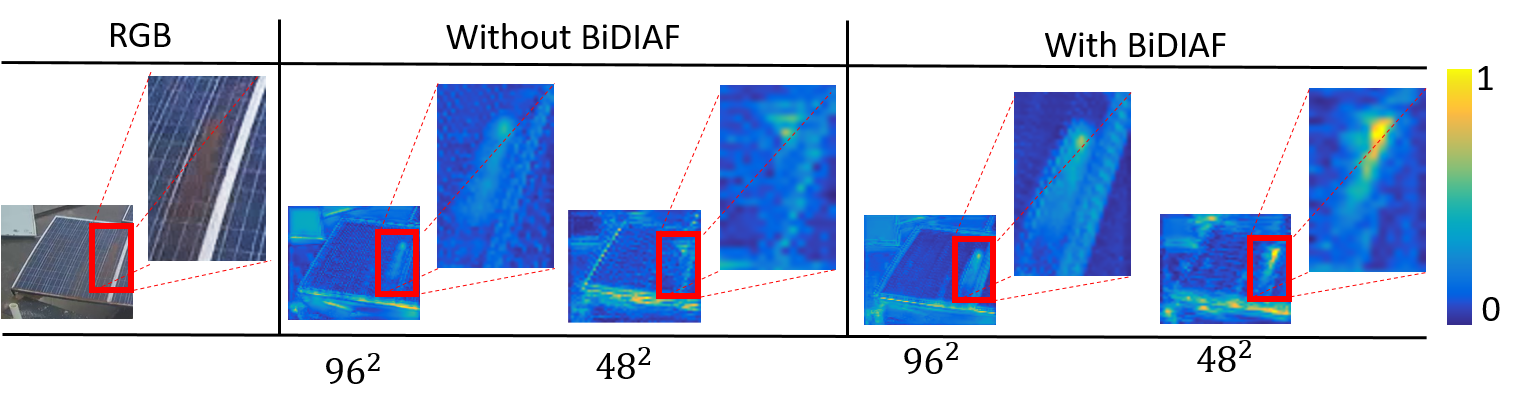}
\caption{Visualization of feature maps with and without BiDIAF. Due to the down-sampling operations, dataset specific features are lost. BiDIAF reinforces the input at different spatial-levels to learn data-specific feature. Best viewed in color.}
\label{fig:resVsResPP}
\end{figure}

\subsection{ImpactNet: Image to Impact Analysis}
Traditional CNNs encode the spatial information about the objects in an image by performing convolution and down-sampling operations in a top-down fashion. Although these CNNs do not encode the localization information, one may combine the feature maps at multiple spatial levels using a bottom-up approach to localize the object. However, the down-sampling operations tend to lose the spatial information and may hinder in object localization (see Figure \ref{fig:resVsResPP}). To address this limitation, we introduce a novel ``bi-directional input-aware fusion block (BiDIAF)'' that reinforces the input inside the network to compensate the loss of spatial information; thereby helping the network learn the relevant features with respect to the input. BiDIAF block takes an input from the main CNN branch ($\mathbf{x}_M$) and shares one of the output with the same branch, hence we call this unit as \textit{bi-directional} input-aware fusion unit.

Our proposed unit can be integrated with any CNN (such as VGG \cite{simonyan2014very} or ResNet \cite{ResNet}). Following the success of ResNet \cite{ResNet} in different visual recognition tasks, we choose ResNet as our baseline network. ResNet stacks residual convolutional units (RCU) to aggregate feature maps at different spatial levels. The input and output in RCU are connected through identity mapping, which improves the information flow inside the network and prevents vanishing gradient issue. We add the BiDIAF unit between two residual convolutional units (RCU), as shown in Figure \ref{fig:resnetPP}. BiDIAF takes the output of previous RCU $\mathbf{x}_R^{l}$ and previous BiDIAF $\mathbf{x}_A^{l}$ (if exists) unit along with an input image $\mathbf{I}$ as input and produces two outputs that are given as input to the next RCU $\mathbf{x}_M^{l+1}$ and next BiDIAF block $\mathbf{x}_A^{l+1}$. We can formulate BiDIAF unit as:
\begin{equation}
\mathbf{x}_A^{l+1} = \mathcal{F}_{B}\left(\left[\mathbf{x}_R^{l}, \ \ \mathcal{F}_{I}\left(\mathbf{I} \right),\ \ \mathcal{F}_{P}\left(\mathbf{x}_A^{l}\right)\right] \right)
\end{equation}
\begin{equation}
\mathbf{x}_M^{l+1} = \mathbf{x}_R^{l} + \mathbf{x}_A^{l+1}
\label{eq:bidiaf}
\end{equation}
The function $\mathcal{F}_{P}$ projects $\mathbf{x}_A^{l}$ to the same dimensionality as $\mathbf{x}_R^{l}$ using  $1\times 1$ convolution. The function $\mathcal{F}_{I}$ first sub-samples $\mathbf{I}$ to the same spatial dimensions as $\mathbf{x}_R^{l}$ using $3\times3$ average pooling operation and then projects the sub-sampled image to the same dimensionality as $\mathbf{x}_R^{l}$ using $5\times 5$ convolution. Apart from projection, $5\times 5$ convolution also learns input relevant feature maps. The function $\mathcal{F}_{B}$ concatenates the feature maps obtained from $\mathbf{x}_R^{l}$, $\mathcal{F}_{I}\left(\mathbf{I} \right)$, and $\mathcal{F}_{P}\left(\mathbf{x}_A^{l}\right)$, followed by $1\times1$ convolution that projects the concatenated feature maps to the same dimensionality as $\mathbf{x}_R^{l}$.

CNNs for classification do not encode the localization information explicitly. Most of the existing methods (e.g. \cite{he2017maskrcnn,hariharan2015hypercolumns,ghiasi2016laplacian, shelhamer2017fully, badrinarayanan2017segnet, noh2015learning, yu2015multi, chen2016deeplab}) use labeled data to learn the localization mask. Data labeling is an expensive task and therefore, we propose a two-fold strategy to generate localization mask: (1) aggregate the feature maps at different spatial-levels to create a candidate mask, and (2) refine the localization masks by jointly training a classification and localization network, assuming candidate masks as ground truth during training.

\subsection{Parameter-free Candidate Mask Creation}
Our approach for localizing the region of impact is motivated by Burt and Adelson's Laplacian pyramid-based method \cite{burt1983laplacian}, which encode and decode the image information using top-down (analysis) and bottom-up (synthesis) pyramids respectively. Standard CNNs aggregate feature maps in top-down fashion; thus suggesting their resemblance with the analysis network. Therefore, we can decode the encoded feature maps using a synthesis pyramid i.e. in bottom-up fashion for localizing the impact area \cite{ghiasi2016laplacian, noh2015learning, badrinarayanan2017segnet}.

\begin{figure}[b!]
\centering
\includegraphics[width=0.8\columnwidth]{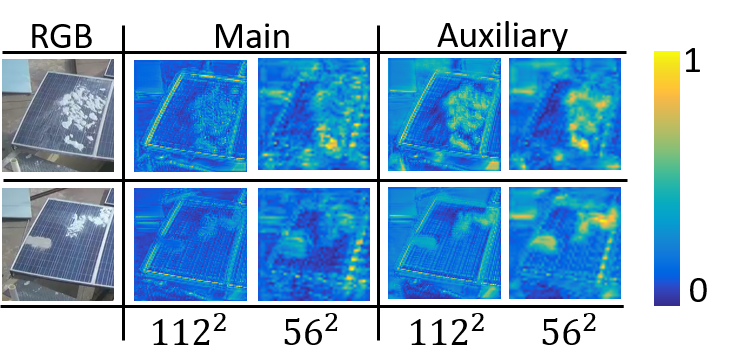}
\setlength{\belowcaptionskip}{-4mm}
\caption{Visualization of feature maps in main and auxiliary branches in the ImpactNet network. For visualization, we have scaled the feature maps to the same scale. Best viewed in color.}
\label{fig:mainAux}
\end{figure}

Our synthesis pyramid fuses the feature maps of main $\mathbf{x}_{M}^l$ and auxiliary $\mathbf{x}_{A}^l$ branches at level $l$ using Eq. \ref{eq:fuseMainAux} to produce a localization mask $\mathbf{x}_{mask}^l$, which is then up-sampled to the same size as level $l-1$ using bilinear interpolation. This process is repeated till the size of localization mask is the same as the input image $\mathbf{I}$. 
\begin{equation}
\mathbf{x}_{mask}^l = (\mathbf{x}_{M}^l \otimes \mathbf{x}_{A}^l) \oplus \mathbf{x}_{A}^l
\label{eq:fuseMainAux}
 \end{equation}
 where $\otimes$ and $\oplus$ are element-wise multiplication and addition operations. The element-wise multiplication operation gives high importance to values only when the feature maps from both the main and auxiliary branches agree; therefore, multiplicative gating helps in suppressing the irrelevant features. The resultant feature map is then combined with the feature map from auxiliary branch using an element-wise addition operation; which boosts the values of relevant features identified using the multiplicative gating. Our inspection of feature maps at different spatial levels reveals that $\mathbf{x}_{A}^l$ has much more descriptive power than $\mathbf{x}_{M}^l$, and therefore, we boost the feature maps using the auxiliary branch $\mathbf{x}_{A}^l$ (see Figure \ref{fig:mainAux}).
 
\noindent \textbf{Mask to label image:} Solar panel image can be split into three classes: background, solar panel, and soiling area. We detect the solar panel by performing Gaussian filtering and edge detection operations on an RGB image. The area outside detected solar panel is assigned a label of 1 (corresponding to background). The area within the detected panel is then thresholded using the mean of the mask. If the pixel value (inside the panel) in the mask is less than the mean, then it is assigned a label of 2 (corresponding to panel). Otherwise, we assign a label of 3 (corresponding to soiling area).

\subsection{Mask FCNN}
Mask FCNN is a fully convolutional CNN that aims to simultaneously predict the soiling impact and the soiling area (localization). Our approach is motivated by the recently proposed method, Mask R-CNN \cite{he2017maskrcnn}, that applies classification and masking in \textit{parallel}. Unlike Mask R-CNN that adopts two stage procedure (region proposal network followed by classification and masking network), our method is fully convolutional i.e. we do not use any region proposal network.

Mask FCNN is composed of two networks: 1) classification network, the ImpactNet, and 2) synthesis network as shown in Figure \ref{fig:resnetPP}. For synthesis network, we follow the fully convolutional bottom-up approach (e.g. \cite{badrinarayanan2017segnet, noh2015learning, ronneberger2015u}). Bottom-up approaches up-sample the feature maps to invert the loss of information due to down-sampling operations. Our bottom-up architecture is a stack of synthesis units (SUs), which can be defined as:

\begin{equation}
\mathbf{x}_S^l = \mathcal{F}_D(\{\mathbf{x}_M^l, \mathbf{x}_A^l,  \mathbf{x}_M^{l+1}, \mathbf{x}_A^{l+1}\})
\end{equation}

$\mathcal{F}_D$ is a composite function comprising of $1\times1$ convolution, $1\times1$ deconvolution, and $3\times3$ convolution operations. $1\times1$ convolution operation reduces the dimension of the feature maps of $\mathbf{x}_M^{l}$ and $\mathbf{x}_A^{l}$ to $C$-dimensional space while $1\times1$ deconvolution up-samples the feature maps of $\mathbf{x}_M^{l+1}$ and $\mathbf{x}_A^{l+1}$ to the same spatial dimensions as $\mathbf{x}_M^{l}$ and $\mathbf{x}_A^{l}$ along with projecting the feature maps to $C$-dimensional space. Mask FCNN was trained by minimizing the multi-task loss $L_{multi}= L_{cls} + L_{mask}$, where $L_{cls}$ and $L_{mask}$ are multinomial cross-entropy loss functions for classification (soiling impact) and masking (soiling area) respectively.

Cleaning actions for solar panels are dependent upon soiling type. For example, potential cleaning actions for bird drop and dust are wiping and air blow respectively. Therefore, it becomes critical to determine the soiling type for efficiently managing the work-force at solar farms. To determine the soiling type, we use a webly supervised classification network (WebNN) and is discussed next.

\subsection{WebNN: Webly Supervised Neural Network}
Given an image of solar panel with soiling mask, WebNN determines the soiling type. WebNN utilizes large amount of web data to train a soiling type classifier and is inspired from \cite{chen2015webly}. We, first, collected images (with and without solar panel) from the Internet by querying the most common soiling categories such as dust (brown, gray, red and black), white chalk powder, bird drop, snow, and crack. These images include soiling categories which were not available in our dataset. 24-dimensional RGB histogram of each of these images were extracted as feature vectors. A small 3-layered neural network, with 50, 100, and 150 hidden neurons per layer respectively, was trained on these feature vectors to predict the soiling type.

To assign a label to the soiling area, we crop the RGB area (referred as ROI in Figure \ref{fig:resnetPP}) corresponding to the soiling area in the localization mask. A 24-dimensional RGB histogram is computed for this ROI as a feature vector, which is then classified using WebNN to predict the soiling type. 

\vspace{-2mm}
\section{Experiments and Results}
\vspace{-2mm}
We performed thorough experiments with various training and model choices on the PV-Net data. We compared classification and localization accuracies of our model, DeepSolarEye. We found that the proposed BiDIAF block improves the classification and localization capabilities of ResNet. We tested the generalizability of our method on images downloaded from the Internet and found that our method was extensible to incorporate the soiling types that were not present in our dataset.

\subsection{Classification Models and Results}
\label{ssec:classRes}
\begin{figure}[b!]
\centering
\begin{subfigure}[b]{\columnwidth}
\centering
\includegraphics[width=0.7\columnwidth]{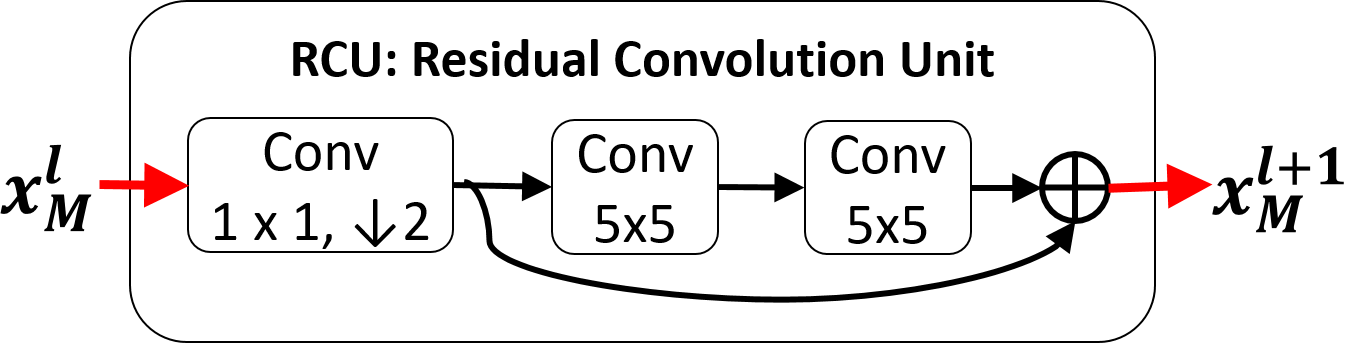}
\caption{Analysis unit in ImpactNet-A}
\label{fig:impactA}
\end{subfigure}
\vfill
\begin{subfigure}[b]{\columnwidth}
\centering
\includegraphics[width=0.9\columnwidth]{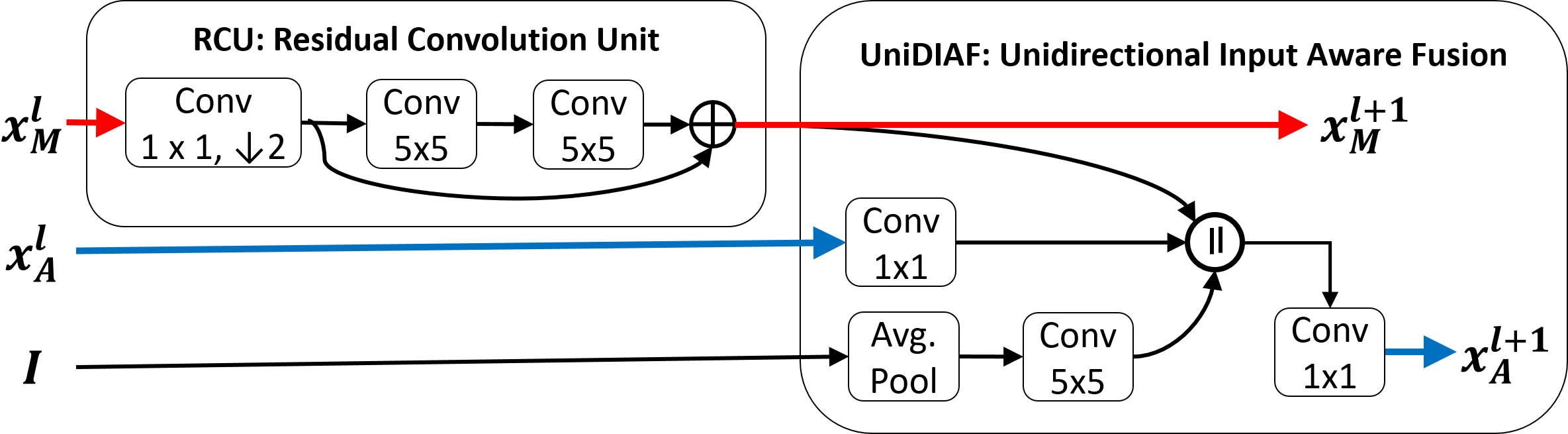}
\setlength{\belowcaptionskip}{-6mm}
\caption{Analysis unit in ImpactNet-B}
\label{fig:impactB}
\end{subfigure}
\setlength{\belowcaptionskip}{-4mm}
\caption{Different types of analysis units used in our experiments. Notations are the same as in Figure \ref{fig:resnetPP}.}
\label{fig:impactTypes}
\end{figure}

\paragraph{Models:} We trained and tested 3 different classification models with single input and single output (SISO) setting (panel image as an input and power loss level as output). Our proposed method (ImpactNet) was compared with its two alternatives. For the first alternative (referred as ImpactNet-A), the BiDIAF block was removed. The resultant network after removing the BiDIAF block is the same as ResNet \cite{ResNet} and has only one CNN branch i.e. main branch (Figure \ref{fig:impactA}). For the second alternative (referred as ImpactNet-B), we modified the BiDIAF Eq. \ref{eq:bidiaf} from $\mathbf{x}_M^{l+1} = \mathbf{x}_R^{l} + \mathbf{x}_A^{l+1}$ to $\mathbf{x}_M^{l+1} = \mathbf{x}_R^{l}$ i.e. input-aware feature maps were not shared with the main branch (Figure \ref{fig:impactB}).

Our model (ImpactNet) was trained and tested  under multiple input and single output (MISO) setting (panel image and environmental factors as inputs and power loss level as output). In our experiments, we used \textit{solar irradiance} and time of the day from the image timestamp as environmental factors. To fuse these multiple inputs, we tried two alternatives: element-wise sum (ImpactNet-C) and concatenation (ImpactNet-D). We emphasize that solar irradiance is an important factor as it captures environmental conditions (such as cloudy and sunny) indirectly, which influences the power loss \cite{ando2015sentinella}.

\vspace{-4mm}
\paragraph{Training:} We trained all of our models end-to-end for 90 epochs using SGD with an initial learning rate of 0.01 decaying it by a factor of 10 after every 30 epochs, momentum of 0.9, weight decay of 0.0005, and a batch size of 32 on a single NVIDIA TitanX GPU. We used spatial dropout \cite{tompson2015efficient} with a dropout probability of 0.2 after every analysis unit (Figure \ref{fig:resnetPP}). The PV-Net dataset (N=45,754) was split randomly into training (N=27,537) and validation (N=18,217) sets. We binned the normalized power loss into $C$ equal bins, with each bin representing a soiling impact level (or class). In our experiments, we varied $C$ from 2 to 16. Inverse class probability weighting scheme was used in loss function to address the class-imbalance issue. Further, we augmented training data using standard augmentation techniques such as horizontal flips, vertical flips, and random rotations. We did not use any color-based augmentation strategies as the color and power generation capacity of the solar panel are directly influenced by the environmental factors, such as sunlight. We would like to highlight that our classification approach was motivated by the solar farm maintenance practices, where maintenance actions were categorized based on severity levels which were captured in power loss bins.

\begin{figure}[b!]
\centering
\includegraphics[width=0.8\columnwidth]{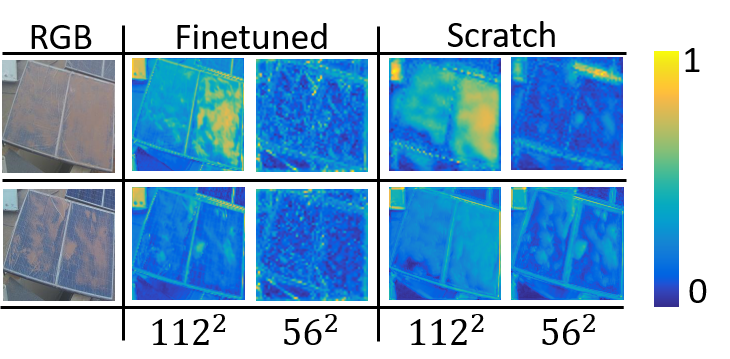}
\setlength{\belowcaptionskip}{-4mm}
\caption{Visualization of feature maps at different spatial resolutions with two different weight initialization strategies. Random weight initialization helped in learning dataset-specific feature maps. For visualization, we scaled the feature maps to the same scale. Best viewed in color.}
\label{fig:featVisScrPre}
\end{figure}

\vspace{-4mm}
\paragraph{Weight initialization:} We trained ResNet-18 on a subset of our dataset (training and validation sets each having 2,000 images) for 4-class classification task, with two different initialization strategies: (1) random weight initialization \cite{he2015delving}, and (2) fine-tuning the model that was trained on the ImageNet. Both networks attained similar accuracy for the 4-class classification task ($95\% \pm 1\%$). Our casual inspection of feature maps revealed that the network initialized with random weights paid attention to the dataset-specific features (see Figure \ref{fig:featVisScrPre}). Therefore, we initialized the network weights randomly.

\vspace{-4mm}
\paragraph{Results:} Classification results were presented in Table \ref{tab:classRes1} and Table \ref{tab:classRes}. We can make the following observations:
\begin{itemize}[itemsep=0pt]
\item[{(1)}] \textit{Convolutional block type:} Replacing the VGG-type blocks with ResNet-type blocks improved the classification accuracy by about 7\%. 
\item[{(2)}] \textit{Effect of BiDIAF:} Replacing the analysis unit in the ImpactNet-A network (Figure \ref{fig:impactA}) with an analysis unit in Figure \ref{fig:impactB} (ImpactNet-B) improved the accuracy by about 1\%. Further, when we replaced the analysis unit in the ImpactNet-A network with the analysis unit in Figure \ref{fig:resnetPP} (RCU + BiDIAF), then the accuracy improved by about 2\% (for both 8- and 16-class networks). The increase in classification accuracy with BiDIAF unit is likely due to the fact that it promotes data-specific feature learning, even at low-spatial resolutions (see Figure \ref{fig:resVsResPP}).
\item[{(3)}] \textit{SISO vs MISO:} By adding environmental factors as input, the accuracy of ImpactNet improved by about $2\%$. The improvement in the accuracy is not drastic; suggesting that ImpactNet was able to learn the complex relationships between environmental factors and soiling that leads to power loss.
\end{itemize}
To further check the performance of our method (ImpactNet-D) in real-world, we tested our method on the data collected using our experimental setup for additional 3 weeks. Our method was able to attain an accuracy of 84.5\% for 8-classes.

\begin{table}[b!]
\centering
\resizebox{0.8\columnwidth}{!}{
\begin{tabular}{l|c|c}
\toprule
& \multicolumn{2}{c}{\textbf{Convolutional Block Type}} \\
\cline{2-3}
& \textbf{VGG-type} \cite{simonyan2014very} & \textbf{ResNet-type} \cite{ResNet} \\
\midrule
w/o BiDIAF & 73.8 & 80.03 \\
w/ BiDIAF & 75.4 & 82.02 \\
\bottomrule
\end{tabular}
}
\setlength{\belowcaptionskip}{-2mm}
\caption{This table compares the top-1 accuracies of different types of convolutional blocks on our dataset (for 8 classes). VGG-type block is the same as ResNet-type block (Figure \ref{fig:impactA}), except the skip connection.}
\label{tab:classRes1}
\end{table}

\begin{table}[b!]
\centering
\resizebox{\columnwidth}{!}{
\begin{tabular}{cl|cccc|c}
\toprule
&\multirow{2}{*}{\textbf{Models}} & \multicolumn{4}{c|}{\textbf{Classes}} & \textbf{\# Params} \\
\cline{3-6}
 &  & \textbf{2} & \textbf{4} & \textbf{8} & \textbf{16} &  (in Million)\\
 \midrule
 \multirow{3}{*}{\rotatebox[origin=c]{90}{SISO}} & \textbf{ImpactNet} & 97.56 & \textbf{93.39} & 82.02 & 68.43 & 1.97 \\
 & \textbf{ImpactNet-A} & 97.77 & 93.24 & 80.03 & 66.68  & 1.96 \\
& \textbf{ImpactNet-B} & 97.61 & 93.18 & 80.99 & 67.88 & 1.97\\
\midrule
\multirow{2}{*}{\rotatebox[origin=c]{90}{MISO}} & \textbf{ImpactNet-C} & 97.64 & 93.10 & 82.97 & 70.19 & 1.99 \\
& \textbf{ImpactNet-D} & \textbf{97.82} & 93.28 & \textbf{83.32} & \textbf{70.59} & 1.99\\
\bottomrule
\end{tabular}
}
\setlength{\belowcaptionskip}{-4mm}
\caption{This table compares the top-1 accuracies of different models on our dataset.}
\label{tab:classRes}
\end{table}

\subsection{Localization Results}
\paragraph{Model and training details:} For localization experiments, we first computed the candidate masks using pyramid-based method, which were then refined using Mask FCNN. We used the  same training and augmentation strategy as discussed in Section \ref{ssec:classRes}. Note that Mask FCNN learned about 2.12 million parameters.

\vspace{-4mm}
\paragraph{Evaluation metrics:} We evaluated the localization performance of our method both subjectively and objectively. For subjective assessment, we computed a \textit{mean opinion score (MOS)}, while for objective assessment, we measured \textit{Jaccard Index (JI)}, a widely used metric for measuring the localization accuracy. 

For measuring the MOS, we divided our entire dataset into non-overlapping interval of 10 minutes and selected an image randomly from every such interval; resulting in 579 images. For each image, we asked following four questions to the user to determine the localization accuracy:
\begin{itemize}[itemsep=0pt]
\item[{\textbf{Q1:}}]  How many regions of dust that were present in the RGB image, but not in the localization mask?
\item[{\textbf{Q2:}}] How many regions of dust that were detected in localization mask, but not present in the RGB image?
\item[{\textbf{Q3:}}] On a scale of 0 to 10, rate the level of under-segmentation with 0 being perfectly segmented and 10 being fully under-segmented. 
\item[{\textbf{Q4:}}] On a scale of 0 to 10, rate the level of over-segmentation with 0 being perfectly segmented and 10 being fully over-segmented.
\end{itemize}
If the area detected in a localization mask was half (or double) of the region of dust in an RGB image, then it was fully under-segmented (or over-segmented).

For measuring the JI, we selected a subset of 241 images out of 579 images. This subset includes all images where we noted high variance in the subjective assessment. These images were then annotated by participants using LabelMe \cite{Russell2008}. We asked participants to annotate the dust regions on the solar panel image along with the dust type (such as brown and gray). These images were used as a ground truth for measuring the localization accuracy (JI).

\begin{figure}[t!]
\centering
    \includegraphics[width=0.75\columnwidth]{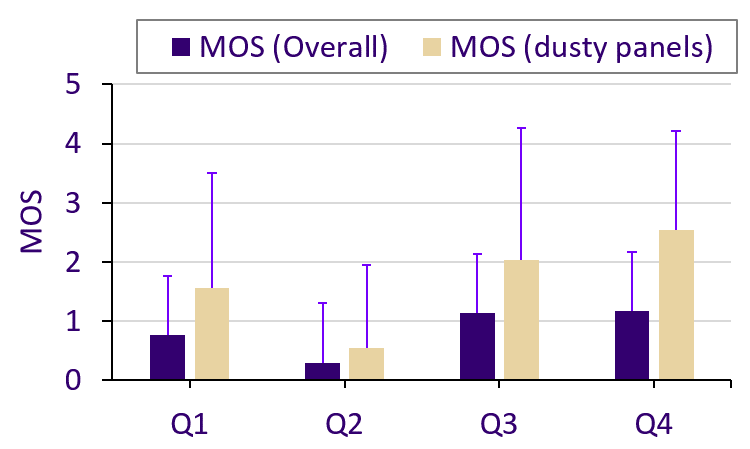}
    \setlength{\belowcaptionskip}{-2mm}
    \caption{Subjective assessment results. Best viewed in color.}
    \label{fig:mos}
\end{figure}

\vspace{-4mm}
\paragraph{Subjective assessment results:} For sufficient cultural, gender, and racial diversity, we used Amazon Mechanical Turk for conducting this experiment. A total of 172 unique users participated in our study, with each image being rated by 5 different users. MOS is shown in Figure \ref{fig:mos}. 
\begin{itemize}[itemsep=0pt]
\item[{(1)}] Across all questions, overall MOS was lower than MOS for soiled (or dusty) panels; suggesting most of the mistakes were made on the soiled panels. However, MOS for soiled images was very low. On soiled images, our method was not able to locate on average 2 soiled patches (Q1) while falsely detecting about 0.5 soiled patches (Q2) per image. 
\item[{(2)}] MOS for under-segmented (Q3) and over-segmented (Q4) images were almost the same and close to 0 (on a scale of 0 to 10); suggesting that over- and under-segmentations were not severe.
\end{itemize}

\vspace{-4mm}
\paragraph{Objective assessment results:} Table \ref{tab:ji} compares the performance of three methods. First two methods used pyramid-based approach for generating the candidate masks while the third method was our end-to-end method, Mask FCNN. For a fair comparison between these models, we masked the background area (non-panel area) and did not consider it while measuring JI. From Table \ref{tab:ji}, we see that BiDIAF unit increased the JI of ImpactNet-A (or ResNet) by about 4\%. This indicates that BiDIAF unit helped in learning input-specific features which resulted in good localization capabilities. Further, Mask FCNN improved the JI of pyramid-based method (with BiDIAF) by about 24\%; suggesting joint learning enabled efficient aggregation of feature maps from the classification network.

\begin{table}[t!]
  \centering
  \resizebox{0.6\columnwidth}{!}{
  \begin{tabular}{|c|c|c|}
  \hline
  \multicolumn{2}{|c|}{\textbf{Method}} & \textbf{JI (in \%)} \\
  \hline
  \textbf{Pyramid-} & \textbf{w/o BiDIAF} & 38 \\
  \cline{2-3}
  \textbf{based} & \textbf{w/ BiDIAF} & 42 \\
  \hline
  \multicolumn{2}{|r|}{\textbf{Mask FCNN}} & \textbf{66} \\
  \hline
  \end{tabular}
  }
  \setlength{\belowcaptionskip}{-4mm}
  \caption{Objective assessment results}
  \label{tab:ji}
\end{table}

\vspace{-4mm}
\paragraph{Webly supervised labeling results:} For measuring the labeling accuracy of WebNN, we used the same 241 images that were annotated by participants along with the type of dust. With webly supervised labeling, we achieved a classification accuracy of about 96.24\%. 

To test the flexibility of WebNN in the wild, we queried the Internet for solar panel images with a soiling type as a keyword (e.g. solar panel image with snow) and downloaded 150 images. The images with low quality (either they were overlayed with text or image resolution was less than $100 \times 100$) were manually discarded. After discarding such images, we were left with 50 images. Out of these 50 images, some of the images had multiple panels and were manually cropped to identify the panel area corresponding to the given keyword. These images were then fed into Mask FCNN to produce the localization mask, which were then classified using WebNN. Our method attained an accuracy of about 87\% on these web images. 

DeepSolarEye was able to learn generalizable representations of different soiling types impacting the power loss. When trained on our dataset and tested on the web images, DeepSolarEye was able to localize the impact area (using Mask FCNN) and classify soiling type (using WebNN) even on the soiling types (e.g. snow, crack, and bird drop) that were not present in our dataset (Figure \ref{fig:siximages}).

\vspace{-2mm}
\section{Application in Solar Farm Maintenance}
\vspace{-2mm}
DeepSolarEye provides enriched information about solar panel soiling and defects (soiling impact, soiling localization, and soiling type). Further, soiling localization could be used to easily compute soiling coverage area. Such information could be used for efficient solar farm monitoring and maintenance. Work force management is one of the crucial task in solar farm management, especially when the farm is spread across several acres. Managing such farm requires to address two main questions: 1) How to clean? and 2) When to clean?

The first question can be answered using the soiling type while the second question can be answered using soiling impact and soiling coverage area (computed from the soiling localization mask). To achieve this, we build a simple decision tree based on the soiling impact, soiling localization, and soiling type. Some results of this decision tree are shown in Figure \ref{fig:siximages}. In Figure \ref{fig:siximages}(c), DeepSolarEye correctly identified soiling type (dust) and suggested correct cleaning type (air blow). Though the soiling impact is low (12.5\% to 25\%) the soiling coverage is about 30\% of the actual panel area. Therefore, the suggested action along with the cleaning priority was \textit{high}. The low impact level was primarily due to the environmental factors (cloudy day). 

\vspace{-2mm}
\section{Conclusion}
\vspace{-2mm}
In this paper, we presented a first CNN-based application for a new domain of solar panel soiling and defect analysis.  Our method, DeepSolarEye, takes an RGB image of a solar panel and environmental factors as inputs and predicts the power loss, soiling localization, and soiling category in real time. We propose a four-stage methodology to train DeepSolarEye in weakly supervised fashion that completely avoids manually labeled localization data. We introduce a novel BiDIAF block for superior localization capabilities. We, further, leveraged the web-crawled data for categorizing the soling type, which allowed inclusion of new soiling types without re-training the Mask FCNN. Our empirical study suggests that BiDIAF module improves the classification and localization capabilities of ResNet (or ImpactNet-A) by about 3\% and 4\%. Our end-to-end model yielded further improvement of about 24\% on localization task when trained in a weakly supervised manner.  Additionally, we constructed a new dataset for solar panel image analysis consisting 45,000+ images. 

Our classification model generalizes well to different domains. We found that the proposed BiDIAF unit improves the classification and localization capabilities of ResNet on the plant disease dataset \cite{sladojevic2016deep} and the Cifar-10/100 dataset \cite{krizhevsky2009learning}. Please see \textbf{Appendix \ref{sec:plantRes} and Appendix \ref{sec:cifar}} for more details.

\vspace{-2mm}
\section*{Acknowledgments}
\vspace{-2mm}
The authors thank Mohit Jain for helping in designing the subjective assessment study. 

{\small
\bibliographystyle{ieee}
\bibliography{main}
}

\clearpage

\appendix

\section{Experimental Set-up}
\label{sec:setup}
\begin{figure}[b!]
	\centering
	\includegraphics[width=0.9\columnwidth]{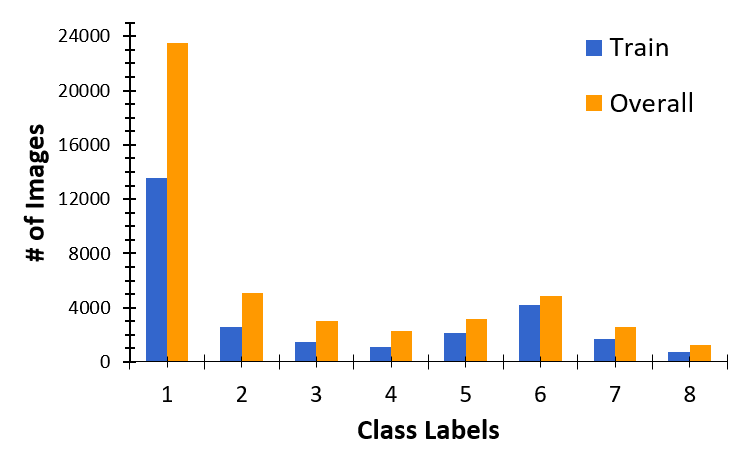}
	\caption{Distribution of class labels for 8-classes}
	\label{fig:distribution}
\end{figure}

We introduce first of its kind dataset, \textit{PV-Net}, comprising of 45,754 images of solar panels with their power loss. We bin the power loss into 8 bins(classes k) of equal sizes. The distribution of data is shown in Figure \ref{fig:distribution}.

\begin{figure}[b!]
	\centering
	\includegraphics[width=0.9\columnwidth]{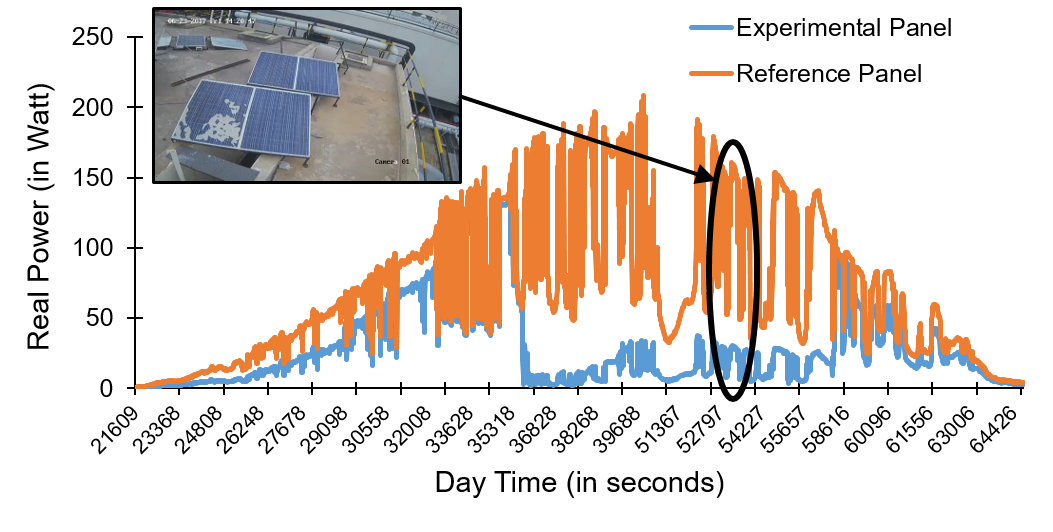}
	\caption{Graph showing the power generated by reference and experimental panel through out the day.}
	\label{fig:powerGen}
\end{figure}

Our experimental setup consists of two identical solar panels, which are kept side by side with an RGB camera facing the panels. Soiling experiments were conducted on the first panel (close to the camera) while the other panel was used for reference purpose. Images were captured at every 5 seconds and corresponding power generated by panels were also recorded (see Figure \ref{fig:powerGen}). Soiling impact is reported as the \textit{percentage power loss} with respect to the reference panel. 

\section{Classification Error Analysis on the PV-Net dataset}
\label{sec:errAna}
\begin{figure}[t!]
	\centering
	\begin{subfigure}{0.8\columnwidth}
		\includegraphics[width=\columnwidth]{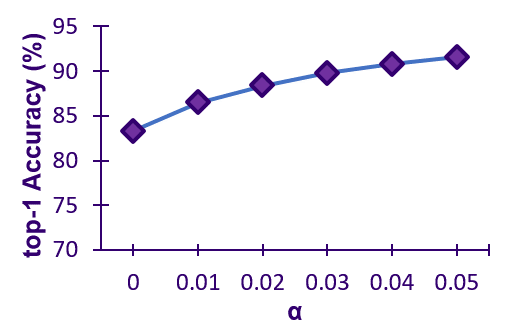}
		\caption{$\alpha$ vs. accuracy}
		\label{fig:ovp}
	\end{subfigure}
	\vfill
	\begin{subfigure}{0.8\columnwidth}
		\includegraphics[width=\columnwidth]{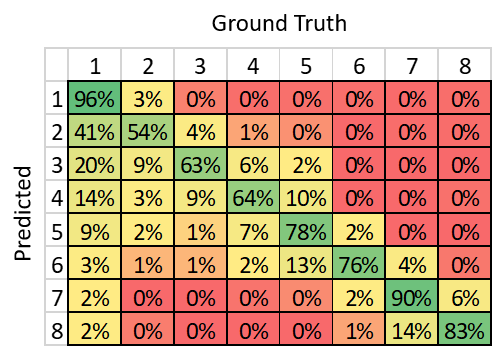} 
		\caption{Confusion Matrix ($\alpha=0$)}
		\label{fig:conmat}
	\end{subfigure}
	\caption{Impact of relaxing the boundary conditions.}
	\label{fig:relaxBound}
\end{figure}

Figure \ref{fig:conmat} visualizes the confusion matrix for 8-class case. We can see that the majority of the mistakes are made with the neighboring classes. Since we binned the classes at fixed interval, it becomes critical to understand the mistakes made by our network i.e. mistakes are made near the boundary or towards the extreme end of the neighboring classes. We introduced a relaxing variable $\alpha$ that relaxes the boundaries of each impact level bin. For example, if the range of the bin is between 12.5\% to 25\%, then after relaxation, the range will be $12.5\% - \alpha$ to $25\% + \alpha$. Figure \ref{fig:ovp} shows a graph between $\alpha$ and accuracy. When we increased $\alpha$ from 0 to 0.01, the accuracy of our method increased from about 83\% to about 88\%; suggesting the mistakes are at the border of the bin and are tolerable.

\section{Experiments with Plant Disease Dataset}
\label{sec:plantRes}
Towards the generalization of our model in other domain for localization task, we experimented with the publicly available plant disease dataset. We trained and tested our model, ImpactNet, with and without BiDIAF for plant disease classification task. In both cases, our model attained an accuracy of around $97\%$, which is comparable to the method proposed by \cite{sladojevic2016deep}. However, on casual visual inspection, we found that ImpactNet without BiDIAF pays more attention to leaf area (such as medrib and veins) while with BiDIAF, it pays more attention to disease area (Figure \ref{fig:plantComp}). This suggests that BiDIAF has promising localization capabilities in other domains, which we intend to explore in detail in future.  
\begin{figure}[t!]
	\centering
	\includegraphics[width=\columnwidth]{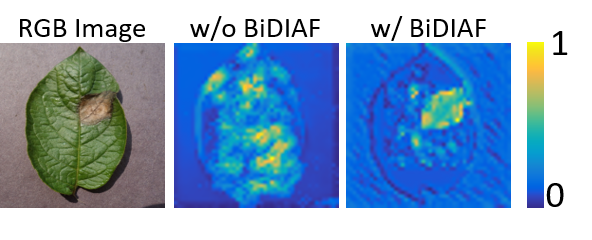}
	\caption{Feature map visualization with and without BiDIAF.}
	\label{fig:plantComp}
\end{figure}

\begin{figure}[t!]
	\centering
	\includegraphics[width=0.9\columnwidth]{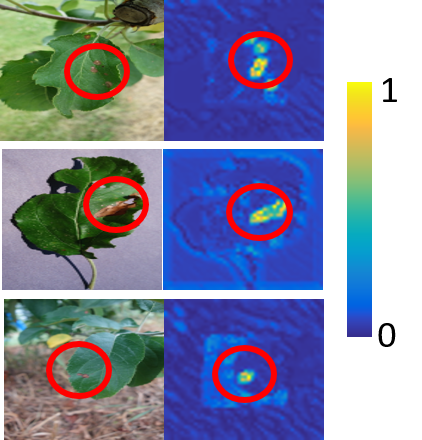}
	\caption{More feature map visualizations with BiDIAF unit on the plant disease dataset. Red circles denote the disease area.}
	\label{fig:plantComp}
\end{figure}

\section{Results on the Cifar Dataset}
\label{sec:cifar}
To show the efficacy of the proposed BiDIAF block, we performed experiments on the Cifar image classification dataset. We trained our model using the same training strategy as in \cite{ResNet}. Experimental results are given in Table \ref{tab:cifarRes}. We observed that the proposed BiDIAF unit improves the accuracy of ResNet by about 1\% and 2.5\% across different depth levels on Cifar-10 and Cifar-100 datasets respectively. 

\begin{table}[t!]
	\resizebox{\columnwidth}{!}{
		\begin{tabular}{c|cc|cc}
			\toprule
			& \multicolumn{2}{|c|}{\textbf{Cifar-10}} & \multicolumn{2}{|c}{\textbf{Cifar-100}}\\
			\cline{2-5}
			\textbf{Depth} & \textbf{ResNet} &  \textbf{ResNet w/ BiDIAF} & \textbf{ResNet} &  \textbf{ResNet w/ BiDIAF}\\
			\midrule
			20  & 91.25  & 92.03 & 71.95  & 73.15\\
			56 & 93.03  & 93.84 &  72.94 & 75.93 \\
			110  & 93.57 & 94.68 & 74.84 & 77.23\\
			\bottomrule
		\end{tabular}
	}
	\caption{This table reports the top-1 test accuracies of ResNet with and without BiDIAF unit on the Cifar-10 dataset. BiDIAF unit establishes a long-range connection between an input image  and any convolutional layer; thereby, promotes learning of dataset-specific features and improves the flow of information inside the network.}
	\label{tab:cifarRes}
\end{table}

\section{Qualitative Results on Solar Panel Images}
\label{sec:moreRes}
\begin{figure*}[!htb]
	\centering
	\begin{minipage}{2\columnwidth}
		\centering
		\includegraphics[width=\columnwidth]{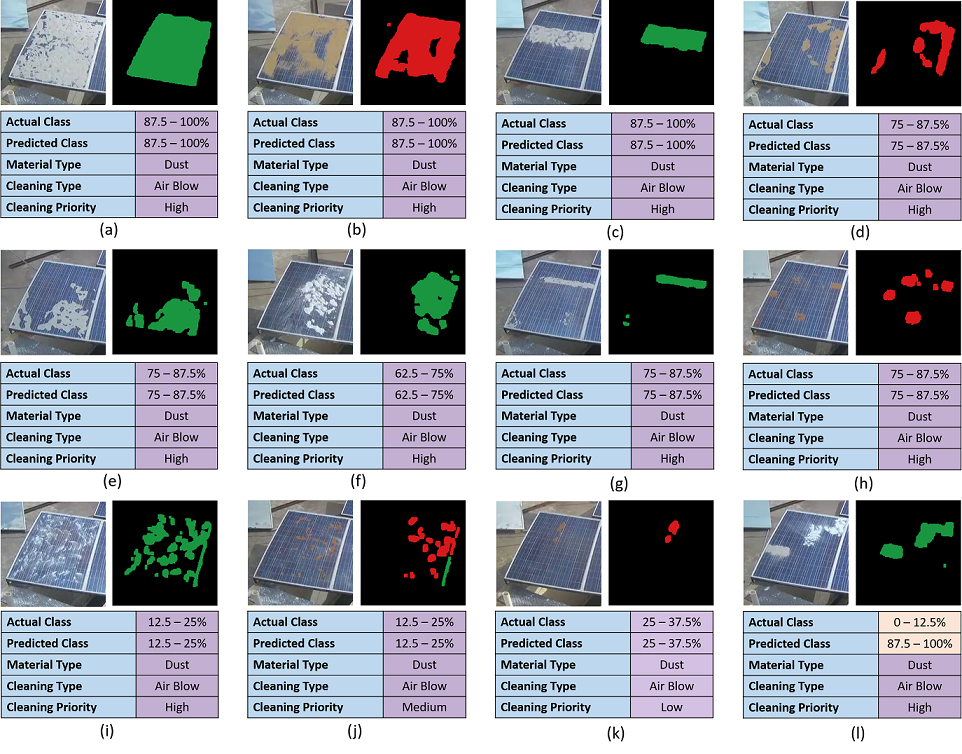}
		\caption{Images (from our dataset) depicting the performance of our method. Best viewed in color.}
		\label{fig:exampleOurDataset}
	\end{minipage}
	\vspace{4mm}
	\begin{minipage}{2\columnwidth}
		\centering
		\includegraphics[width=\columnwidth]{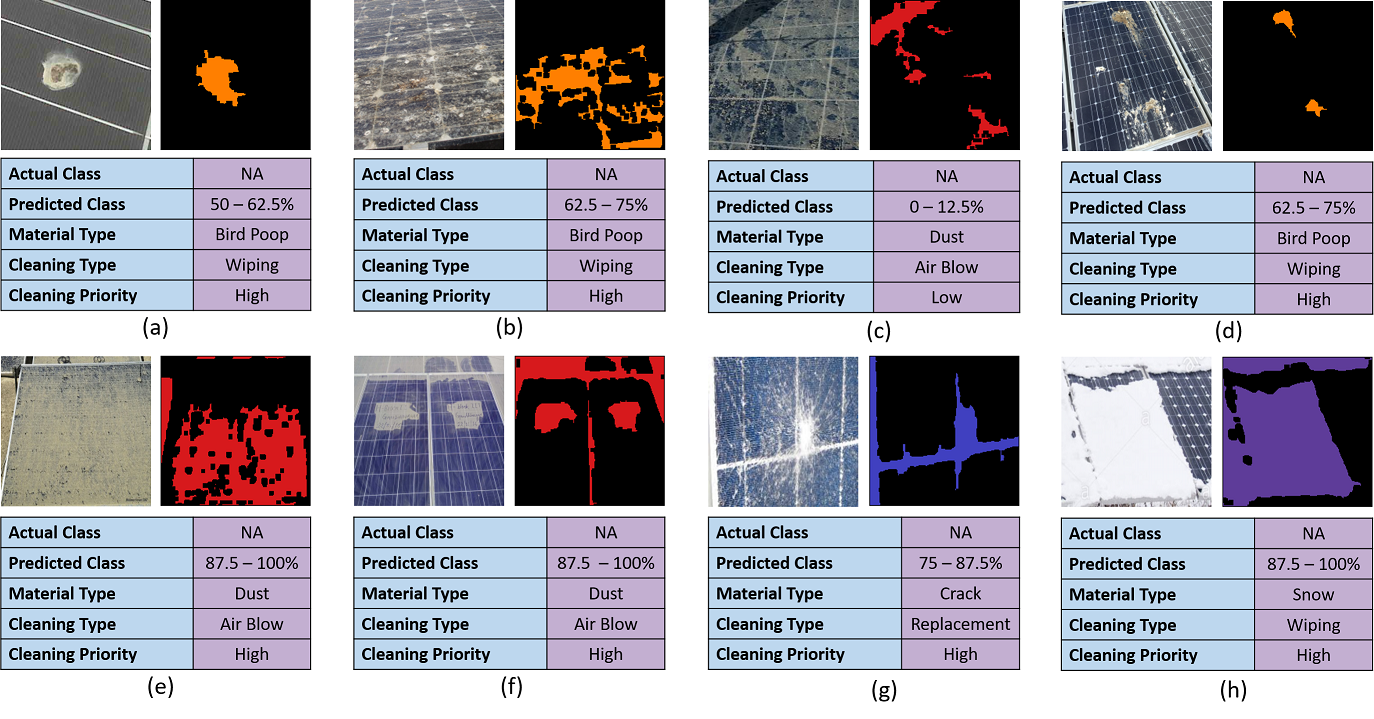}
		\caption{Images (from the Internet) depicting the performance of our method in the wild. Best viewed in color.}
		\label{fig:exampleWeb}
	\end{minipage}
\end{figure*}

Figure \ref{fig:exampleOurDataset} and \ref{fig:exampleWeb} depicts the performance of our method on different solar panel images (from our dataset as well as the Internet). We can see that DeepSolarEye has good localization and soiling classification properties.

\end{document}